%% file: root.tex
\documentclass[letterpaper, 10 pt, conference]{ieeeconf}  

\IEEEoverridecommandlockouts                              

\usepackage{graphics} 
\usepackage{epsfig} 
\usepackage{times} 
\usepackage{amsmath} 
\usepackage{amssymb}  
\usepackage{amsfonts}
\usepackage{bm}
\usepackage{ulem}
\usepackage{algorithm}
\usepackage{algorithmic}
\usepackage{booktabs}
\usepackage[flushleft]{threeparttable}
\usepackage{multirow}
\usepackage{hyperref}
\usepackage{xcolor}
\usepackage{svg}

\hypersetup{
    colorlinks=true,
    linkcolor=blue,
    urlcolor=blue,
    citecolor=blue,
    filecolor=blue,
    pdfborder={0 0 0}
}

\title{\LARGE \bf
LiLoc: Lifelong Localization using Adaptive Submap Joining \\ and Egocentric Factor Graph
}

\author{Yixin Fang$^{1,2}$, Yanyan Li$^{3,4}$, Kun Qian$^{2\dagger}$, Federico Tombari$^{4,5}$, Yue Wang$^{1}$, Gim Hee Lee$^{3}$
\thanks{$^{1}$ Zhejiang University; $^{2}$ Southeast University; $^{3}$ National University of Singapore; $^{4}$ Technical University of Munich; $^{5}$ Google Zurich.}%
\thanks{$\dagger$ Corresponding author: Kun Qian, Email: kqian@seu.edu.cn.}%
}

\begin{document}

\maketitle
\thispagestyle{empty}
\pagestyle{empty}

\begin{abstract}
This paper proposes a versatile graph-based lifelong localization framework, \textit{LiLoc}, which enhances its timeliness by maintaining a single central session while improves the accuracy through multi-modal factors between the central and subsidiary sessions. First, an adaptive submap joining strategy is employed to generate prior submaps (keyframes and poses) for the central session, and to provide priors for subsidiaries when constraints are needed for robust localization. Next, a coarse-to-fine pose initialization for subsidiary sessions is performed using vertical recognition and ICP refinement in the global coordinate frame. To elevate the accuracy of subsequent localization, we propose an egocentric factor graph (EFG) module that integrates the IMU preintegration, LiDAR odometry and scan match factors in a joint optimization manner. Specifically, the scan match factors are constructed by a novel propagation model that efficiently distributes the prior constrains as edges to the relevant prior pose nodes, weighted by noises based on keyframe registration errors. Additionally, the framework supports flexible switching between two modes: relocalization (RLM) and incremental localization (ILM) based on the proposed overlap-based mechanism to select or update the prior submaps from central session. The proposed \textit{LiLoc} is tested on public and custom datasets, demonstrating accurate localization performance against state-of-the-art methods. Our codes will be publicly available on \url{https://github.com/Yixin-F/LiLoc}.
\end{abstract}

\vspace{0.5em}


\section{Introduction}
\begin{figure}[!t]
    \centering
    \includegraphics[width=\linewidth]{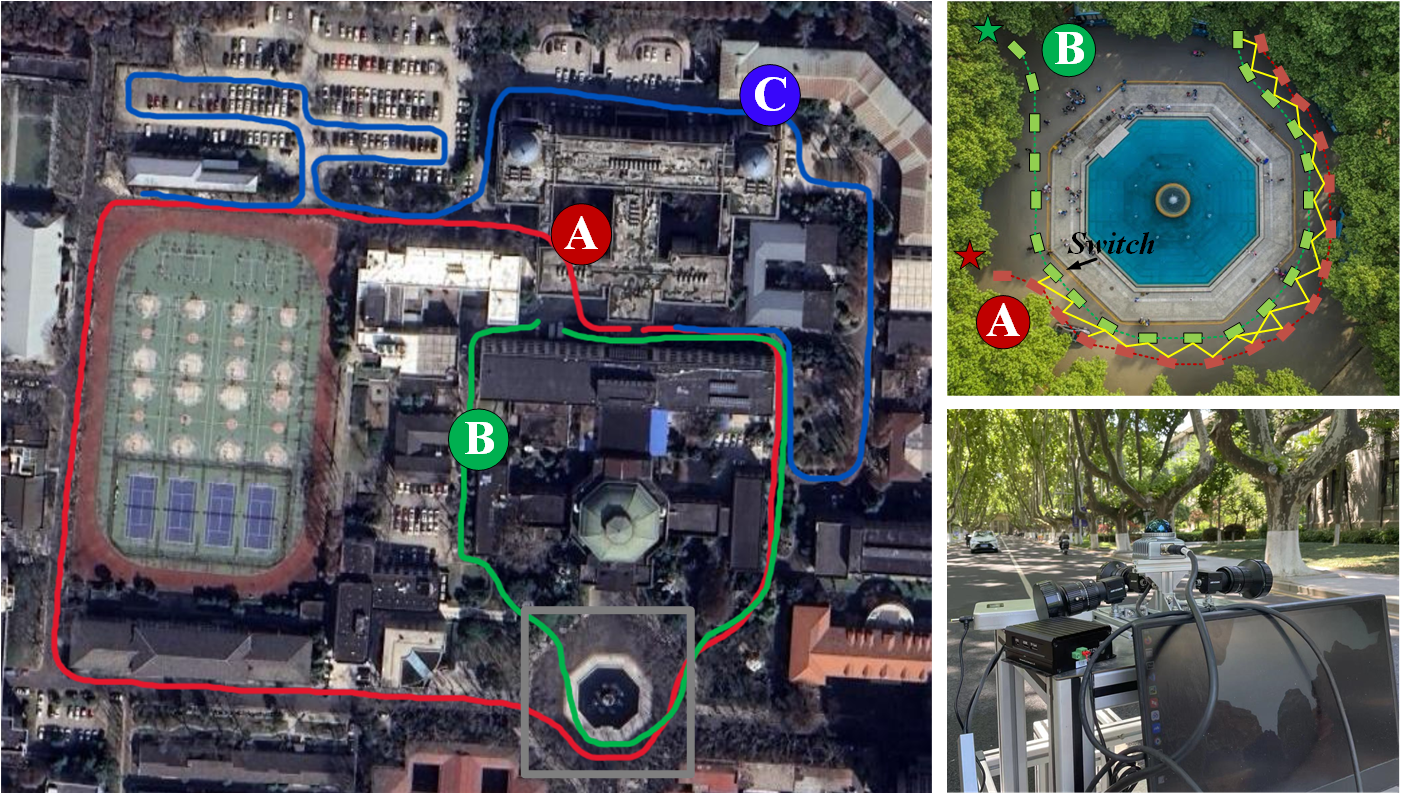}
    \caption{A real-world application of LiLoc with central session A and subsidiary sessions B and C. In the top-right subfigure, the pentagons represent the start points, rectangles denote covered nodes, dashed lines indicate odometry constraints, and solid lines represent prior constraints. Since Session B passes through the area covered by Session A, it switches from ILM to RLM. The bottom-right shows our data collection equipment.}
    \label{fig:show}
\end{figure}

Accurate localization with long-term effectiveness is key for robots to achieve autonomous navigation while interacting with changing environments, particularly in large-scale, underground and industrial exploration scenarios. Lifelong localization can be framed as a problem of multi-session mapping, where a given central session (e.g., session A in Fig. \ref{fig:show}) is continually leveraged and updated, while achieving long-term localization for other subsidiary sessions (e.g., sessions B and C in Fig. \ref{fig:show}) \cite{kim2022lt}. Although general SLAM algorithms such as LiDAR-based \cite{zhang2014loam, shan2020lio, xu2022fast, fang2024segmented} and structure-based \cite{li2021rgb, li2016manhattan} perform well in single-session mapping, it is challenging to extend them to long-term localization due to their limited capacity to handle spatial and temporal correlations in multi-session environments.

A number of sensors and related algorithms can provide reliable poses for a robot's localization in a given map. Hardware-assisted localization with GNSS/INS can directly obtain the current position of the robot in the global coordinate frame; however, due to signal blockages in certain scenarios, it cannot always provide real-time localization information \cite{gu2015gnss}. Map-based localization \cite{egger2018posemap, koide2019portable} often replaces or corrects the motion prediction from the odometry module with the relative pose obtained by aligning current scan to prior maps, which can directly constrain current pose estimation using prior knowledge. However, they fail to account for the discrepancies between current scan and prior maps due to environmental changes, making the relative poses less reliable and causing drift in the overall pose estimation. Recently, \cite{feng2024block, koide2024tightly} have been presented to improve the effectiveness of prior constrains by converting the relative pose between current scan and prior maps into scan matching factor and adding it into the factor graph of odometry. While they can yield relatively reliable pose estimates through factor graph optimization (FGO), it still overlooks the errors arising from point cloud registration, risking over-relying on prior constraints. Moreover, by focusing solely on the odometry-based FGO, these approaches lead to potential misuse of prior information and loss of accuracy.

\begin{figure*}[!t]
    \centering
    \includegraphics[width=7.0in]{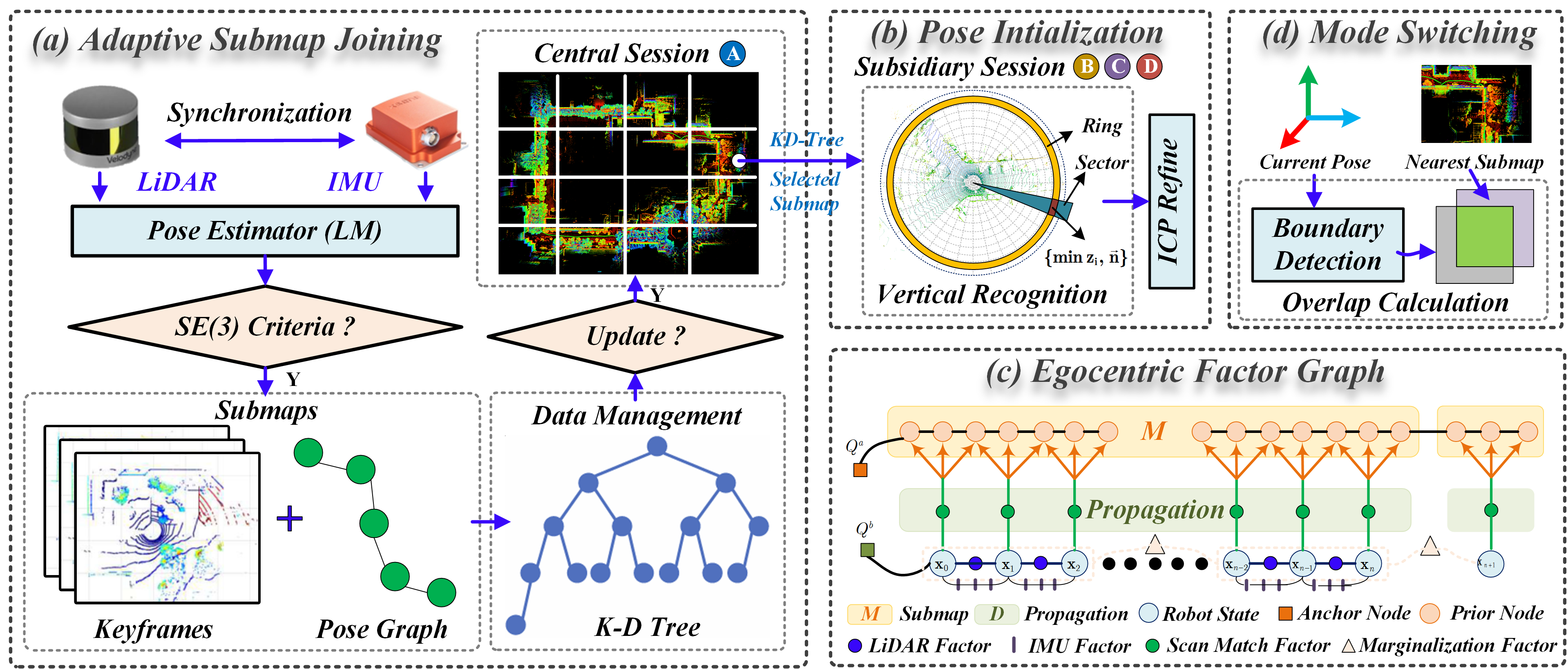}
    \caption{System overview of our proposed LiLoc, which consists of four main modules: (a) adaptive submap joining, (b) pose initialization, (c) egocentric factor graph (EFG), and (d) mode switching.}
    \label{fig:system}
\end{figure*}

To address these limitations, we propose a versatile graph-based framework for long-term localization tasks with two flexible localization modes, namely RLM and ILM. The carefully designed framework consists of four modules: adaptive submap joining, pose initialization, EFG and mode switching, as depicted in Fig. \ref{fig:system}. The adaptive submap joining strategy is introduced to enable the interaction between the central and subsidiary sessions. It dynamically manages (i.e., generates, selects and updates) prior submaps from the central session, reducing memory consumption by minimizing the maintenance of large-scale map while preserving the timeliness of prior knowledge. The pose initialization provides accurate initial poses as prior values for subsidiary sessions, supporting subsequent graph-based estimation. The EFG, built in a centralized manner, tightly integrates IMU preintegration, LiDAR odometry and scan matching factors. Specifically, in handling scan matching factors, the EFG introduces an innovative propagation model. Compared to other factor graph-assisted methods \cite{feng2024block, koide2024tightly}, the propagation model distributes the relative pose between current scan and prior maps as edges to relevant prior pose nodes, weighted by noises based on keyframe registration errors. By leveraging the joint optimization of prior and current factor graphs, this approach not only preserves the constraining influence of prior constraints on pose estimation but also alleviates the detrimental effects of their uncertainty on accuracy. Meanwhile, the overlap-based mode-switching mechanism allows flexible localization modes, making the system effective for multi-session tasks. 
Our contributions are as follows:
\begin{itemize}
    \item A graph-based framework for long-term localization featuring a flexible mode-switching mechanism, to achieve accurate multi-session localization.
    
    \item An adaptive submap joining strategy to dynamically manage (i.e., generate, select and update) prior submaps, reducing system memory consumption while ensuring the timeliness of prior knowledge.
    
    \item An egocentric factor graph (EFG) module to tightly couple multi-modal constrains, along with a novel propagation model to enhance  prior constrains by distributing weighted scan matching factors in joint factor-graph optimization (JFGO). 
    
    \item We achieve the competitive performance on public and custom datasets and the proposed system will be released for community use.
\end{itemize}

\section{Related Work}

 For 3D map building, the LOAM system \cite{zhang2014loam} uses planar and edge features from LiDAR scans to estimate a robot's pose, minimizing drift while balancing accuracy and efficiency by separating odometry and mapping. Building on LOAM, LIO-SAM \cite{shan2020lio} tightly integrates IMU data with LiDAR through FGO in a sliding window, improving drift correction and state estimation over time in challenging terrains. FAST-LIO2 \cite{xu2022fast} further enhances state estimation with the Iterated Error State Kalman Filter (IESKF), which iteratively refines poses by updating feature correspondences, reducing errors and ensuring reliable pose estimation.

Global localization often relies on Monte Carlo Localization (MCL) \cite{thrun2001robust}, which is widely used in robotic systems \cite{chen2021range, sun2020localising} for its robustness against local minima and ability to handle ambiguities, converging particle sets to a consensus through iterative measurements and motion updates for reliable localization. For global optimization tasks, Branch-and-Bound Search (BBS) \cite{hess2016real, feng2024block} is highly effective, systematically exploring the search space and pruning non-optimal regions to reduce computational costs while maintaining accuracy. This makes BBS well-suited for autonomous driving, where precision and reliability are crucial in dynamic environments.

Map-based localization methods, such as HDL-Loc \cite{koide2019portable} and Fastlio-Loc adapted from FAST-LIO2 \cite{xu2022fast}, have developed SLAM-type relocalization frameworks that incorporate standalone odometry modules, which directly use additional scan-to-map information as constraints to refine odometry estimation. To improve the accuracy of localization, block-map-based localization (Block-Loc) \cite{feng2024block} utilizes a factor graph that integrates the IMU preintegration factor and scan matching factor to obtain optimized LiDAR poses. It also leverages pre-constructed block maps (BMs) to reduce the computational burdens of maintaining large-scale maps. Alternatively, a range-inertial localization algorithm \cite{koide2024tightly} is proposed to tightly integrate scan-to-scan and scan-to-map point cloud registration with IMU data in a sliding window factor graph framework. However, these methods do not perform robust and reasonable processing of scan matching constrains for efficient utilization.


\section{Problem Formulation}
Let $\bm{\mathcal{W}}$ denote the world coordinate frame. The transformation from the body coordinate (IMU frame) $\bm{\mathcal{I}}$ to $\bm{\mathcal{W}}$ at timestamp $t$ is represented as $\bm{T}_t = \begin{bmatrix}\bm{R}_t & \bm{P}_t \\ \bm{0} & 1 \end{bmatrix} \in SE(3)$, where $\bm{R}_t$ is the rotation matrix and $\bm{P}_t$ is the translation vector. We assume that a single session $\bm{\mathcal{S}}$ contains pose graph $\bm{\mathcal{G}=(\mathcal{V}, \ \mathcal{E})}$, where $\bm{\mathcal{V}}=\{\bm{T}_1, \ \bm{T}_2, \ ..., \  \bm{T}_n\}$ are the pose nodes and $\bm{\mathcal{E}}=\{\bm{E}_1, \ \bm{E}_2, \ ..., \  \bm{E}_n\}$ are the constraint edges, and keyframe database $\bm{\mathcal{F}} = \{\bm{F}_1, \ \bm{F}_2, \ ..., \ \bm{F}_n\}$. 

Our goal is to continuously estimate LiDAR poses for subsidiary sessions, such as B and C, based on the prior knowledge from the central session A. The adaptive submap joining strategy is first employed to generate prior submaps for $\bm{\mathcal{S}}^{a}$, and allows for selection or update of these submaps when necessary. For a subsidiary session  B represented by $\bm{\mathcal{S}}^{b}$, after obtaining an initial pose $\bm{T}^b_{init}$ in the global coordinate frame of $\bm{\mathcal{S}}^{a}$, the subsequent pose estimation using the EFG can be formulated as a Maximum A Posteriori (MAP) problem, which aims to minimize the overall error in the joint factor graph $\{\bm{\mathcal{G}}^a, \bm{\mathcal{G}}^b\}$:
\begin{equation}
    \arg \min_{\bm{\mathcal{G}}^b} \sum_{i\in \bm{\mathcal{E}}^{\bm{I}_b}} \bm{e}^{\bm{I}_b}_i + \sum_{j\in \bm{\mathcal{E}}^{\bm{L}_b}} \bm{e}^{\bm{L}_b}_j + \sum_{k\in \bm{\mathcal{E}}^{a}} \bm{W}(\bm{e}^{a}_k)
    \label{eq:all}
\end{equation}
where $\bm{\mathcal{E}}^{\bm{I}_b}$ and $\bm{\mathcal{E}}^{\bm{L}_b}$ represent the IMU preintegration and LiDAR odometry edges in subsidiary session B, respectively. $\bm{\mathcal{E}}^{a}$ denotes the scan matching edges between the central session A and subsidiary session B. The corresponding error terms, or factors, $\bm{e}^{\bm{I}_b}$, $\bm{e}^{\bm{L}_b}$ and $\bm{e}^{a}$, are tightly coupled through the EFG with a propagation model. Before optimizing the joint factor graph $\{\bm{\mathcal{G}}^a, \bm{\mathcal{G}}^b\}$, each graph is initialized with its anchor node with prior noise \cite{kim2010multiple}. The mode-switching function $\bm{W}(\cdot)$, valued between zero (ILM) and one (RLM), controls the integration of $\bm{e}^{a}$. Further details are provided in Section \ref{s:ego}. The following sections will illustrate the long-term localization for subsidiary session B using prior knowledge from central session A as an example.

\section{Methodology}

\subsection{Adaptive Submap Joining}
\label{s:submap}
The proposed submap joining strategy is integrated into the entire system to dynamically manage (i.e., generate, select, and update) prior submaps from central session A, ensuring spatial continuity in the prior constraints for smooth and stable pose estimation in subsidiary session B. Submap generation is initially implemented to build the database of submaps within the prior knowledge, based on the distance along the \textit{SE}(3) manifold of odometry poses. Assuming the maximum number of keyframes in a submap is $N_s = 20$, and the last pose node in the previous submap $\bm{M}_j$ is $\bm{T}_j$, the \textit{SE}(3) distance $\bm{d}_{kj}$ between the current pose $\bm{T}_k$ and $\bm{T}_j$ is computed as $\bm{T}_{kj}$. The generation criteria is defined as:
\begin{equation}
    \bm{M}_k = \bigcup_{i \in (j, k]} \bm{T_i} \cdot \bm{F_i} , \ \text{if } k-j \ge N_s \text{ or } \bm{P}_{kj} \ge h_{p} 
\label{eq:generate}
\end{equation}
where $\bm{M}_k$ denotes the newly generated submap and $h_{p} = 20$ m represents the threshold of translation.

The operations of selecting and updating submaps depend on the two mutually exclusive modes, namely RLM and ILM, during the pose estimation process. As shown in Fig. \ref{fig:system}, the prior poses and the centroids of prior submaps are managed by kd-trees, $Tree_p$ and $Tree_m$, respectively. For each new pose $\bm{T}_{k}^b$ from subsidiary session B, if the mode is RLM, we search for its nearest prior poses in $Tree_p$. If at least one nearest neighbor is found, $\bm{T}_{k}^b$ is registered to $Tree_p$ and associated with the nearest prior submap $\bm{M}_k^a$ in $Tree_m$. Conversely, if the mode is ILM, a new submap is generated to update the database of prior submaps according to Eq. (\ref{eq:generate}). Further details are provided in Algorithm \ref{alg:1}.

\subsection{Coarse-to-Fine Pose Initialization}

Pose initialization in the central session $\bm{\mathcal{S}}^{a}$ is utilized to initialize the subsequent pose estimation for subsidiary session B by providing an initial prior pose and associated noise for the factor graph $\bm{\mathcal{G}^b}$. Given a coarse pose estimate $\bm{T}_{co}^b$ for the first keyframe $\bm{F}_{co}^b$, the adaptive submap joining strategy is employed to select the most relevant prior submap $\bm{M}_{co}^a$ from the submaps of central session A. This selection process ensures that the prior information is spatially aligned and suitable for accurate pose refinement. Inspired by  \cite{fang2024segmented} and \cite{kim2018scan}, we transfer each point $\bm{\delta}_k=\{x_k, y_k, z_k\}$ in $\bm{F}_{co}^b$ to a polar coordinate $\bm{\sigma}_k=\{\rho_k, \theta_k, z_k\}$, where $\rho_k=\sqrt{x_k^2+y_k^2}$ and $\theta_k=arctan(y_k/x_k)$. Then, $\bm{F}_{co}^b$ is segmented by equally dividing the polar coordinates in azimuthal and radial directions into $N_a = 60$ sectors and $N_r = 20$ rings. Each segment $\bm{S}_{ij}$ is represented by:

\begin{equation}
    \begin{aligned}
    \bm{S}_{ij}=\{(\vec{\bm{n}}_{ij}, \max_{\bm{\sigma}_k \in \bm{S}_{ij}} z_k) |  \frac{i\cdot L_{max}}{N_r} \le \rho_k < \frac{(i+1)\cdot L_{max}}{N_r}, \\ \frac{j\cdot 2\pi}{N_a} \le \theta_k < \frac{(j+1)\cdot 2\pi}{N_a} \}
    \end{aligned}
    \label{eq:sc}
\end{equation}
where $\vec{\bm{n}}_{ij}$ denotes the normal direction of $\bm{S}_{ij}$ and $L_{max}$ represents the maximum range of LiDAR. 

Similarly, we use Eq. (\ref{eq:sc}) to process all keyframes in the prior submap $\bm{M}_{co}^a$ and generate a candidate set $\bm{V}_{co}^a$. The best candidate keyframe $\bm{F}_{best}^a$ is then selected from $\bm{V}_{co}^a$ based on a similarity measure:

\begin{equation}
    \bm{F}_{best}^a = \min_{\bm{F}_t^a \in \bm{M}_{co}^a} \mathbb{XOR}(\bm{F}_{co}^b, \bm{F}_t^a)
    \label{eq:init}
\end{equation}
where $\mathbb{XOR}(\cdot)$ denotes the function for matrix difference calculation, which can be detailly refer to \cite{kim2018scan}. Then, ICP \cite{rusinkiewicz2001efficient} is used to get the refined initial pose $\bm{T}_{init}^b$.

\subsection{Egocentric Factor Graph (EFG)}
\label{s:ego}

\textit{1) IMU Preintegration Factor:} We use the IMU preintegration technique \cite{forster2016manifold} to efficiently incorporate IMU constraints into the factor graph $\bm{\mathcal{G}}^b$. The predicted motion comprising rotation $\bm{R}$, position $\bm{P}$ and velocity $\bm{V}$ is integrated from IMU measurements over time. The IMU preintegration factor $\bm{e}^{\bm{I}_b}_{ij}$ captures the system's evolution between two time steps $i$ and $j$, providing the relative motion in terms of rotation $\Delta\bm{R}_{ij}$, position $\Delta\bm{P}_{ij}$ and velocity $\Delta\bm{V}_{ij}$:

\begin{equation}
    \begin{aligned}
        \bm{e}^{\bm{I_b}}_{ij} &= ||\mathrm{log}(\bm{R}^T_{i}\bm{R}_{j})||^2 \\
                &+ ||\bm{R}^T_{i}(\bm{P}_j-\bm{P}_i-\bm{V}_i\Delta t_{ij}-\frac{1}{2}\bm{g}\Delta t^2_{ij})||^2 \\
                &+ ||\bm{R}^T_{i}(\bm{V}_j-\bm{V}_i-\bm{g}\Delta t_{ij})||^2
    \end{aligned}
    \label{eq:imu}
\end{equation}
Here, the IMU biases $\bm{b}^a$ and $\bm{b}^\omega$ are jointly optimized alongside the LiDAR odometry factors in the graph.

\textit{2) LiDAR Odometry Factor:} When a new LiDAR scan arrives, the point cloud is undistorted by linear interpolation using IMU preintegration data, and the LiDAR points are incrementally managed by an ikd-tree \cite{cai2021ikd}. For a newly saved keyframe $\bm{F}_{k}^b$, we search for the five nearest points in the ikd-tree based on the initial pose provided by IMU preintegration. The optimal transformation ${\bm{T}_{k}^b}$ is then computed using the Levenberg-Marquardt (LM) algorithm to minimize the distance between each new point and its corresponding local plane patch, which can be formulated as follows:

\begin{equation}
    \bm{T}_{k}^b = \min_{\bm{T}_{k}^b} \sum_{i \in F_{k}^b} \omega_{i} \frac{\begin{vmatrix} (\bm{p}_{k, i} - \bm{p}_{k-1, u}) \\ (\bm{p}_{k-1, u} - \bm{p}_{k-1, v}) \times (\bm{p}_{k-1, u} - \bm{p}_{k-1, w}) \end{vmatrix}}
        {|(\bm{p}_{k-1, u} - \bm{p}_{k-1, v}) \times (\bm{p}_{k-1, u} - \bm{p}_{k-1, w})|}
    \label{eq:lidar}
\end{equation}
where $i$, $u$, $v$ and $w$ are the indices in their corresponding sets, $\omega_{i}$ are the weight according to roughness. Then the LiDAR odometry factor can be formulated as:

\begin{equation}
    \bm{e}^{L_b}={(\bm{T}_{k-1}^b)}^T \cdot \bm{T}_{k}^b
    \label{eq:lidar2}
\end{equation}


\begin{algorithm}[!t]
\caption{LiLoc using EPG}
\label{alg:1}
\begin{algorithmic}[1]
\REQUIRE Prior knowledge $\bm{\mathcal{S}}^a$ from central session A, input LiDAR and IMU data, a coarse initial pose $\bm{T}^b_{co}$ for subsidiary session B;

\STATE Generate prior submaps by the adaptive submap joining strategy and construct kd-trees $Tree_{p}$ and $Tree_{m}$;
\STATE Obtain the initial pose $\bm{T}^b_{init}$ through Eq. (\ref{eq:init}) with ICP;
\STATE Implement IMU preintegration and add this factor to local EFG through Eq. (\ref{eq:imu});
\STATE Estimate LiDAR odometry pose $\bm{T}_k^b$ through Eq. (\ref{eq:lidar}) and add this factor to local EFG through Eq. (\ref{eq:lidar2});
\STATE Add $\bm{T}_k^b$ as a new pose node to graph $\bm{\mathcal{G}}^b$;

\IF{ O($\bm{T}_k^b$) $\ge h_o$ through Eq. (\ref{eq:overlap})}
    \STATE Turn into Relocalization Mode (RLM);
    \IF{$Search(Tree_{m}, \bm{T}_k^b) == true$}
        \STATE Select the nearest prior submap $\bm{M}^a_k$;
        \STATE Get neighbor prior nodes $\{\bm{T}_{n1}^a, \bm{T}_{n2}^a, \bm{T}_{n3}^a\}$ by $Search(Tree_{p}, \bm{T}_k^b)$;
        \STATE Calculate scan match factors and add them to local EFG through Eq. (\ref{eq:match});
    \ENDIF
\ELSE
    \STATE Turn into Incremental Localization Mode (ILM);
    \IF{ $(\bm{T}_{k-1}^b)^{-1} \cdot \bm{T}_{k}^b$ satisfy Eq. (\ref{eq:generate})}
        \STATE Update prior submaps by adding a new submap;
    \ENDIF
\ENDIF
\STATE Implement JFGO for $\{\bm{\mathcal{G}}^a, \bm{\mathcal{G}}^b\}$ through Eq. (\ref{eq:all});
\IF{sliding window changes} 
    \STATE Marginalization through Eq. (\ref{eq:marg});
\ENDIF
\RETURN Optimized pose $(\bm{T}_k^b)^{*}$;
\end{algorithmic}
\end{algorithm}

\textit{3) Scan Match Factor:} To constrain the relative transformation between the current LiDAR scan and the selected prior submap $\bm{M}^a_k$, we utilize Normal Distributions Transform (NDT) \cite{biber2003normal} based registration errors. Unlike recent approaches \cite{feng2024block, koide2024tightly} that use scan match factors by directly incorporating the relative pose as LiDAR odometry edges into $\bm{\mathcal{G}}^b$, we propose a propagation model (illustrated in Fig. \ref{fig:system}) to distribute the relative pose across neighboring prior nodes in $\bm{M}^a_k$, weighted by noises based on registration errors. This approach establishes more reliable edges between the joint graph $\{\bm{\mathcal{G}}^a, \bm{\mathcal{G}}^b\}$, thereby strengthening the constraints of prior knowledge on current pose estimation.

Given the NDT-based transformation $\bm{T}^a_r$ of current scan in $\bm{\mathcal{W}}$, we use the $Tree_p$ structure in Session \ref{s:submap} to search for the $N_d = 3$ nearest neighbor prior nodes $\{\bm{T}_{n1}^a, \bm{T}_{n2}^a, \bm{T}_{n3}^a\}$. Additionally, we compute the registration fitness scores between the current scan $\bm{F}_k^b$ and each of the prior keyframes ${\bm{F}_{n1}^a, \bm{F}_{n2}^a, \bm{F}_{n3}^a}$. Let $\Omega_o$ denote the base noise matrix. The scan match factors can then be formulated as:

\begin{equation}
    \bm{e}^a = \bigcup_{i \in \{ n1, n2, n3\} } || ((\bm{T}^a_{i})^{-1} \bm{T}^a_r) ||_{\Omega_o \bm{f}_b^{i}}
    \label{eq:match}
\end{equation}
where $\bm{f}_b^{i}$ denotes the fitness score. To balance the joint optimization between $\bm{\mathcal{G}}_a$ and $\bm{\mathcal{G}}_b$, we set anchor node $\bm{Q}^a$ with smaller prior noise than $\bm{Q}^b$ \cite{kim2010multiple}. 

\textit{4) Marginalization:} We introduce a marginalization strategy for $\bm{\mathcal{S}}^b$, with a variable sliding window size based on the pose estimation mode. For the RLM, we retain $N_m^r = 5$ keyframes and marginalize older variables by linearizing their factors, resulting in the following linear system:

\begin{equation}
    \begin{bmatrix}
        \bm{H}_{\alpha\alpha} & \bm{H}_{\alpha\beta} \\
        \bm{H}_{\beta\alpha} & \bm{H}_{\beta\beta}
    \end{bmatrix}
    \begin{bmatrix}
        \bm{x}_{\alpha} \\
        \bm{x}_{\beta}
    \end{bmatrix}
    =
    \begin{bmatrix}
        \bm{b}_{\alpha} \\
        \bm{b}_{\beta}
    \end{bmatrix}
    \label{eq:marg}
\end{equation}
where $\beta$ denotes the variables to be marginalized, while the dependent variables are represented as $\alpha$. Applying the Schur complement results in a linear system $\hat{\bm{H}}_{\alpha\alpha} \bm{x}_{\alpha} = \hat{\bm{b}}_{\alpha}$ with:

\begin{equation}
    \begin{aligned}
        \hat{\bm{H}}_{\alpha\alpha} &= \bm{H}_{\alpha\alpha} - \bm{H}_{\alpha\beta}\bm{H}_{\beta\beta}^{-1}\bm{H}_{\beta\alpha} \\
        \hat{\bm{b}}_{\alpha} &= \bm{b}_{\alpha}-\bm{H}_{\alpha\beta}\bm{H}_{\beta\beta}^{-1}\bm{b}_{\beta}
    \end{aligned}.
\end{equation}
When the localization mode switches to the ILM, the marginalization process is triggered. We retain $N_m^l = 10$ keyframes in the current sliding window, and all preceding factors are marginalized as prior information.

\subsection{Mode-Switching Mechanism}
\label{s:mode}
The mode-switching module is performed by the overlap connection between current scan and the nearest prior submap, enabling a flexible alteration between RLM and ILM. As shown in Fig. \ref{fig:system}, for the current pose $\bm{T}_k^b$, a square boundary with side length $2l$ is calculated. This process of overlap connection is formulated as:
\begin{equation}
    \bm{O}(\bm{T}_k^b) = [\bm{P}_k^b(x) \pm l, \bm{P}_k^b(y) \pm l] \ \cap [\bm{\mathcal{B}}(x), \bm{\mathcal{B}}(y)]
    \label{eq:overlap}
\end{equation}
where $\bm{O}(\cdot)$ denotes the overlap ratio and $\bm{\mathcal{B}}(\cdot)$ represents the xy- boundary of selected prior submap. Given a threshold $h_o = 0.7$, the modes are switched as shown in Algorithm \ref{alg:1}.

\section{Experiments}

\subsection{Datasets}
We evaluate our method on two public datasets, NCLT \cite{carlevaris2016university} and M2DGR \cite{yin2021m2dgr}, as well as our custom dataset. The NCLT dataset includes a 10 Hz Velodyne HDL-32E LiDAR and a 100 Hz 9-axis IMU, while the M2DGR dataset uses a Velodyne VLP-32 LiDAR and a 150 Hz 9-axis IMU. Our custom dataset is recorded with a Livox MID 360 and its built-in IMU (details in Fig. \ref{fig:show}). All experiments are conducted on an AMD R9-7940HS CPU with 48GB RAM.

\begin{table}[!t]
\centering
\setlength{\tabcolsep}{5pt}
\caption{Comparison of Global Pose Initialization }
\label{tab:pi}
\begin{threeparttable}
\begin{tabular}{l | c c c c}
\toprule 
Method & xyz (m) & rpy (rad) & F-Score & time (s)  \\
\midrule 
FPFH + RANSAC \cite{buch2013pose}  & 0.33 & 0.12 & 0.24 & 56.36 \\

FPFH + Teaser \cite{yang2020teaser}  & 0.19 & 0.10 & 0.17 & 35.50 \\

FPFH + Quatro \cite{lim2022single} & 0.14 & 0.11 & 0.16 & 31.62 \\

BBS$^{*}$ \cite{hess2016real} & 5.00  & -- & -- & 9.82 \\

BBS++$^{*}$ \cite{feng2024block} & \textbf{0.05} & -- & -- & \textbf{0.87} \\

Ours & 0.07 & \textbf{0.08} & \textbf{0.12} & 1.63 \\
\bottomrule 
\end{tabular}
\begin{tablenotes}
\footnotesize
\item[1] $^{*}$ denotes the experimental results taken from \cite{hess2016real} and \cite{feng2024block}.
\end{tablenotes}
\end{threeparttable}
\end{table}

\subsection{Pose Initialization}
To evaluate the accuracy of pose initialization, we measured the absolute 6-DoF error (``xyz" and ``rpy"), ICP fitness score (F-Score) \cite{rusinkiewicz2001efficient}, and time cost of various methods, including RANSAC-based \cite{buch2013pose}, Teaser-based \cite{yang2020teaser}, Quatro-based \cite{lim2022single}, BBS \cite{hess2016real}, and BBS++ \cite{feng2024block}. Methods from \cite{buch2013pose}, \cite{yang2020teaser}, and \cite{lim2022single} require significant time (over an hour) to initialize map descriptors with FPFH \cite{rusu2009fast}, making them unsuitable for global localization. Compared to BBS \cite{hess2016real}, BBS++ \cite{feng2024block} employs a greedy strategy to enhance localization efficiency. Our proposed coarse-to-fine initialization uses vertical recognition in the selected submap followed by ICP optimization, achieving a better balance between accuracy and time. Results are summarized in Table \ref{tab:pi}.

\begin{table*}[!t]
\centering
\setlength{\tabcolsep}{4.5pt}
\caption{Absolute Trajectory Errors (RMSE, Meters) and Time Cost (Millisecond) Comparison on Public Single-Session Sequences}
\label{table:single}
\begin{threeparttable}
\begin{tabular}{l ccc ccc ccc ccc ccc}
\toprule
\multirow{2}{*}{Method} & \multicolumn{3}{c}{\textit{nclt\_120115}} & \multicolumn{3}{c}{\textit{nclt\_120122}} & \multicolumn{3}{c}{\textit{nclt\_120202}} & \multicolumn{3}{c}{\textit{m2dgr\_s1}} & \multicolumn{3}{c}{\textit{m2dgr\_s2}} \\
\cmidrule(r){2-16}
 & xyz & xyz\_rpy & time & xyz & xyz\_rpy & time & xyz & xyz\_rpy & time & xyz & xyz\_rpy & time & xyz & xyz\_rpy & time \\
\midrule
LIO-SAM \cite{shan2020lio} & $\times$ & $\times$ & $\times$ & $\times$ & $\times$ & $\times$ & $\times$ & $\times$ & $\times$ & 1.451 & 3.102 & 25.33 & 6.303 & 6.877 & 24.16 \\
FAST-LIO2 \cite{xu2022fast} & \underline{1.856} &\underline{3.382} & 34.11 & \underline{1.840} & 3.374 & \underline{29.30} & 1.845 & \underline{3.381} & \underline{21.36} & \underline{0.419} & 2.777 & 22.48 &  \underline{2.848} & 3.955 & 21.02 \\
Ours (ILM) & 2.079 & 4.120 & \underline{33.19} & 2.633 & \underline{3.271} & 30.40 & \underline{1.796} & 3.701 & 26.73 & 1.245 & \underline{2.364} & \underline{20.61} & 3.171 & \underline{3.905} & \underline{19.41} \\ 
\midrule
Hdl-Loc \cite{koide2019portable} & $\times$ & $\times$ & $\times$ & $\times$ & $\times$ & $\times$ & $\times$ & $\times$ &$\times$ & $\times$ & $\times$ & $\times$ & $\times$  & $\times$ & $\times$ \\
Fastlio-Loc & $\times$ & $\times$ & $\times$ & 3.019 & 4.137 & \textbf{32.22} & 0.179 & 2.834 & \textbf{26.18} & 0.145 & 2.752 & \textbf{14.75} & \textbf{0.130} & 2.751 & \textbf{11.68} \\
Block-Loc \cite{feng2024block} & 0.973 & 2.990 & \textbf{28.68} & \textbf{0.198} & 2.834 & 31.72 & 0.185 & 2.833 & 32.15 & 0.233 & 2.756 & 24.66 & 0.220 & 2.757 & 25.40\\
Ours (RLM) & \textbf{0.726} & \textbf{2.948} & 38.99 & 0.199 & \textbf{2.204} & 44.08 & \textbf{0.174} & \textbf{2.491} & 42.47 & \textbf{0.135} & \textbf{1.983} & 38.50 & 0.144 & \textbf{1.631} & 36.77 \\
\bottomrule
\end{tabular}
\begin{tablenotes}
\footnotesize
\item[1] $\times$ denotes that the system totally failed.
\end{tablenotes}
\label{table:comparison}
\end{threeparttable}
\end{table*}

\begin{figure}[!t]
    \centering
    \includegraphics[width=3.4in]{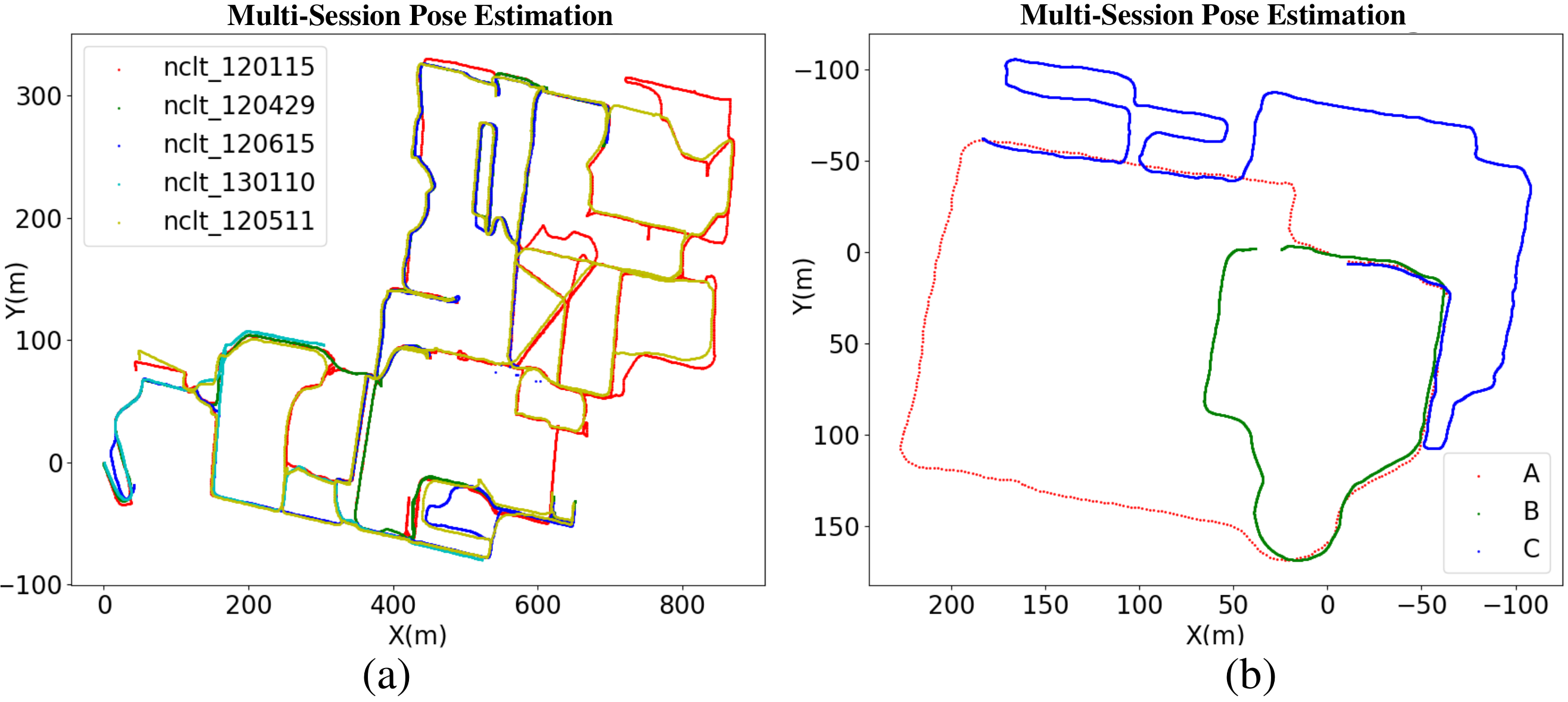}
    \caption{Merged Trajectories resulted from multi-session pose estimation on (a) NCLT dataset and (b) our custom dataset.}
    \label{fig:nclt_multi}
\end{figure}

\begin{figure*}[!t]
    \centering
    \includegraphics[width=6.2in]{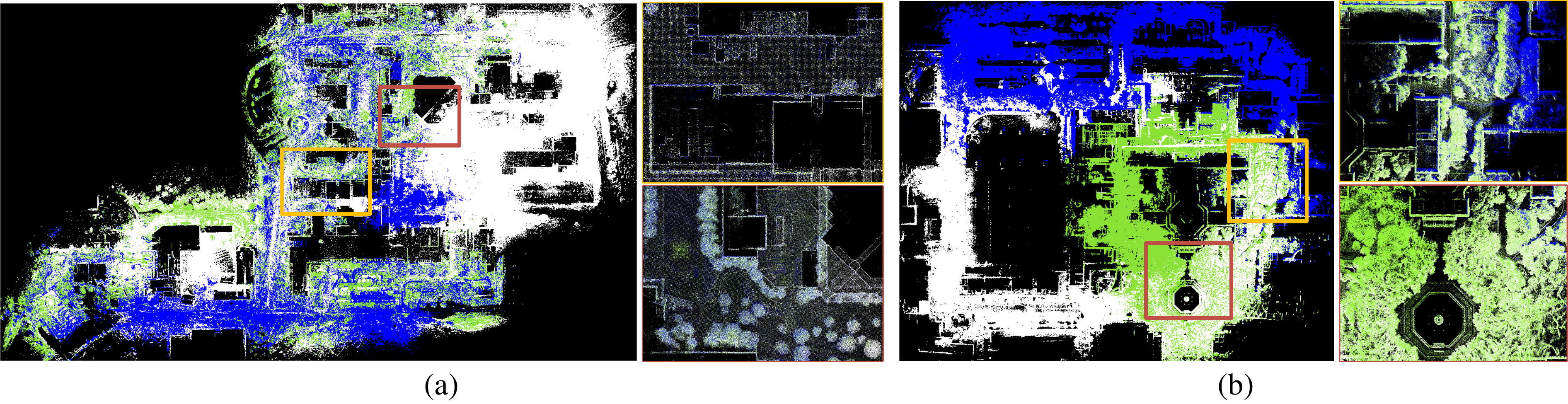}
    \caption{Merged maps from (a) the NCLT dataset and (b) our custom dataset, with the right subfigures showing mapping details to indicate pose estimation accuracy. (a) \textit{nclt\_120115} (white) is used as the central session, with localization for subsidiary sessions \textit{nclt\_120429} (green) and \textit{nclt\_120615} (blue). (b) The same experiment is conducted using central session A (white) and subsidiary sessions B (green) and C (blue).}
    \label{fig:school}
\end{figure*}


\begin{table*}[!t]
\centering
\setlength{\tabcolsep}{4.5pt}
\caption{Absolute Trajectory Errors (RMSE, Meters) and Time Cost (Millisecond) Comparison on Public Multi-Session Sequences}
\label{table:multi}
\begin{threeparttable}
\begin{tabular}{l ccc ccc ccc ccc}
\toprule
\multirow{3}{*}{Method} 
&\multicolumn{12}{c}{\textit{nclt\_120115}} \\ 
\cmidrule(r){2-13}
& \multicolumn{3}{c}{\textit{nclt\_120429}} & \multicolumn{3}{c}{\textit{nclt\_120615}} & \multicolumn{3}{c}{\textit{nclt\_120511}} & \multicolumn{3}{c}{\textit{nclt\_130110}} \\
\cmidrule(r){1-13}
 & xyz & xyz\_rpy & time & xyz & xyz\_rpy & time & xyz & xyz\_rpy & time & xyz & xyz\_rpy & time \\
Hdl-Loc \cite{koide2019portable} & $\times$ & $\times$ & $\times$ & $\times$ & $\times$ & $\times$ & $\times$ & $\times$ & $\times$ & $\times$ & $\times$ & $\times$\\
Fastlio-Loc & $\times$ & $\times$ & $\times$ & $\times$ & $\times$ & $\times$ & 8.246 & 12.774  & 33.41 & 5.447 & 8.620 & \textbf{27.31} \\
Block-Loc \cite{feng2024block} & $\times$ & $\times$ & $\times$ & $\times$ & $\times$ & $\times$ & $\times$ & $\times$ & $\times$ & 9.691 & 11.258 & 37.59\\
Ours w/. 6459 Nodes & 1.692 & \textbf{3.020} & \textbf{47.90} & 2.146 & 4.912 & 48.08 & 1.680 & \textbf{3.674} & 42.44 & \textbf{1.232} & 2.881 & 47.26 \\
\midrule
Ours w/. 8994 Nodes & \textbf{1.378} & 3.307 & 51.47 & \textbf{2.047} & \textbf{4.789} & 53.59 & \textbf{1.409} &  3.806 & 49.54 & 1.377 & \textbf{2.702} & 50.51 \\
Ours w/. 4218 Nodes & $\times$ & $\times$ & $\times$ & 3.489 &  6.102 & \textbf{35.11} & 3.202 & 5.197 & \textbf{32.64} & 2.486 & 4.732 & 32.47 \\
\bottomrule
\end{tabular}
\begin{tablenotes}
\footnotesize
\item[1] $\times$ denotes that the system totally failed. \ \ $^\text{2}$ $\text{w/.}$ denotes the number of active prior nodes.
\end{tablenotes}
\label{table:comparison}
\end{threeparttable}
\end{table*}


\subsection{Single-Session Pose Estimation}

We conducted a single-session evaluation to compare two SLAM algorithms LIO-SAM \cite{shan2020lio} and FAST-LIO2 \cite{xu2022fast}, as well as three localization algorithms HDL-Loc \cite{koide2019portable}, Fastlio-Loc and Block-Loc \cite{feng2024block}. For the ILM evaluation (shown in the first part of Table \ref{table:single}), FAST-LIO2, which utilizes the Iterated Error State Kalman Filter (IEKF) for state estimation, demonstrates stable and efficient ``xyz" and ``xyz\_rpy" performance. In comparison, LIO-SAM, which also relies on a FGO backend, is outperformed by our proposed LiLoc in all aspects. This is primarily because LiLoc employs an ikd-tree \cite{cai2021ikd} to ensure consistent scan-to-map registration and reduce processing time. For the RLM evaluation (shown in the second part of Table \ref{table:single}), our proposed LiLoc significantly surpasses other algorithms in ``xyz" and ``xyz\_rpy" accuracy. This improvement is attributed to the EFG with propagation model that effectively distributes prior constraints to neighboring nodes via the propagation model, thereby creating more edges between joint factor graphs. However, the enhanced accuracy in RLM comes with additional computational time, as the joint optimization inevitably increases computational burdens. The detailed analysis of the time costs can be referred to Section \ref{s:time}.

\subsection{Multi-Session Pose Estimation}

We conducted a multi-session evaluation with HDL-Loc \cite{koide2019portable}, Fastlio-Loc and Block-Loc \cite{feng2024block} on the NCLT dataset \cite{carlevaris2016university} and our custom dataset. The results demonstrate that our proposed LiLoc framework significantly outperforms the other methods in both ``xyz" and ``xyz\_rpy" pose estimation accuracy, as shown in the first part of Table \ref{table:multi} for the NCLT dataset.
In this comparison, HDL-Loc and Block-Loc are unable to switch to the ILM, leading to almost complete failure in multi-session scenarios. On the other hand, Fastlio-Loc uses the ICP \cite{rusinkiewicz2001efficient} algorithm to refine poses generated by FAST-LIO2, which makes it less time-consuming but also more prone to instability. In contrast, our LiLoc framework leverages a robust EFG module and supports autonomous mode switching between ILM and RLM, making it more effective and reliable for multi-session pose estimation tasks. Fig. \ref{fig:nclt_multi} and Fig. \ref{fig:school} present the final merged trajectories and point-cloud maps in the global coordinate frame, which indirectly indicate the accuracy of our long-term localization.

\subsection{Time Cost Evaluation}
\label{s:time}

\begin{figure}[!t]
    \centering
    \includegraphics[width=3.1in]{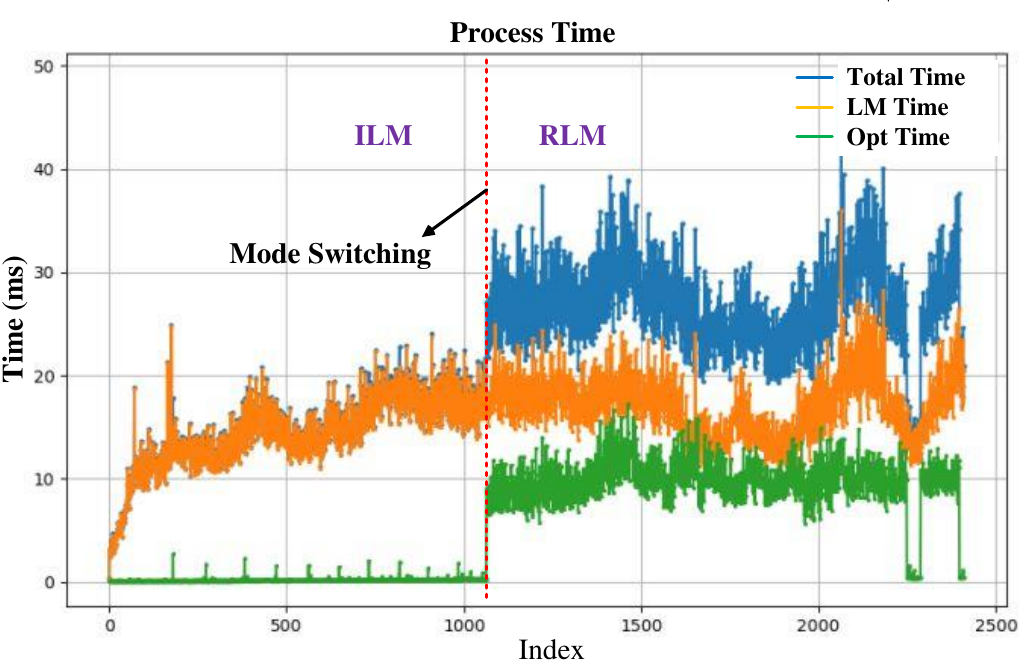}
    \caption{Time cost evaluation of two processes: the LM solver (LM Time) and the JFGO using EFG (Opt Time).}
    \label{fig:time}
\end{figure}

In this section, we evaluate the time cost of the LM solver and JFGO using EFG. As shown in Fig. \ref{fig:time}, the JFGO takes about 10-20 ms, making it slightly more time-consuming than the other methods listed in Table \ref{table:single} and Table \ref{table:multi}. Additionally, we conducted ablation experiments to analyze the impact of the number of prior pose nodes and the results are presented in the second part of Table \ref{table:multi}.

\section{Conclusions}
In this paper, we have proposed a versatile graph-based framework for long-term localization tasks with a flexible mode-switching mechanism, named LiLoc. To efficiently manage the prior knowledge, we proposed an adaptive submap joining strategy, which enables the interaction between the central and subsidiary sessions. Furthermore, we proposed an egocentric factor graph (EFG) with a novel propagation model that effectively distributes prior constraints to relevant prior nodes, weighted by noises based on registration errors, excelling on both single-session and multi-session pose estimation. In the future, we plan to extend our system to multi-robot exploration in unknown environments.





\normalem
\input{root.bbl}

\bibliographystyle{IEEEtran}

\end{document}